\documentclass[a4paper, 10pt, conference]{ieeeconf}      
\usepackage{FG2024}

\usepackage{amsmath}

\DeclareMathOperator*{\argmin}{arg\,min}

\usepackage{booktabs}

\FGfinalcopy 
\IEEEoverridecommandlockouts                              
\usepackage{graphics} 
\usepackage{epsfig} 
\usepackage{mathptmx} 
\usepackage{times} 
\usepackage{amsmath} 
\usepackage{amssymb}  
\usepackage{xcolor}
\usepackage{lipsum}

\def\FGPaperID{****} 

\title{\LARGE \bf
    Learning to Simplify Spatial-Temporal Graphs in Gait Analysis
}

\author{\parbox{16cm}{\centering
    {\large Adrian Cosma and Emilian Radoi}\\
    {\tt\small cosma.i.adrian@gmail.com, emilian.radoi@upb.ro}\\
    {\normalsize
    University Politehnica of Bucharest\\
    }}
}

\begin{document}

\ifFGfinal
\thispagestyle{empty}
\pagestyle{empty}
\else
\author{Anonymous FG2024 submission\\ Paper ID \FGPaperID \\}
\pagestyle{plain}
\fi
\maketitle

\begin{abstract}

 Gait analysis leverages unique walking patterns for person identification and assessment across multiple domains. Among the methods used for gait analysis, skeleton-based approaches have shown promise due to their robust and interpretable features. However, these methods often rely on hand-crafted spatial-temporal graphs that are based on human anatomy disregarding the particularities of the dataset and task. This paper proposes a novel method to simplify the spatial-temporal graph representation for gait-based gender estimation, improving interpretability without losing performance. Our approach employs two models, an upstream and a downstream model, that can adjust the adjacency matrix for each walking instance, thereby removing the fixed nature of the graph. By employing the Straight-Through Gumbel-Softmax trick, our model is trainable end-to-end. We demonstrate the effectiveness of our approach on the CASIA-B dataset for gait-based gender estimation. The resulting graphs are interpretable and differ qualitatively from fixed graphs used in existing models. Our research contributes to enhancing the explainability and task-specific adaptability of gait recognition, promoting more efficient and reliable gait-based biometrics.
\end{abstract}

\section{Introduction}
Gait analysis \cite{gait-survey} utilizes the unique patterns of individuals' walking styles as distinctive identifiers. This non-invasive method of biometric processing has found significant applications across several domains, including security and surveillance \cite{cosma22gaitformer} and healthcare \cite{costilla2020deep}. Gait patterns offer insights into the identity of a person \cite{cosma22gaitformer}, demographics \cite{Catruna2021FromFT}, emotions \cite{emotion_estimation:Randhavane} and mental state \cite{cosma2023psymo}.

Generally, gait analysis approaches have relied on skeleton-based \cite{teepe2021gaitgraph} or silhouette-based methods \cite{gait_set:Chao}. Skeleton-based methods, where the human body is represented as a set of interconnected joints \cite{alphapose}, offer several advantages as they are less sensitive to changes in clothing and lighting conditions, and they provide a more robust and interpretable feature set \cite{teepe2022towards}. Current state-of-the-art methods for skeleton-based gait analysis \cite{teepe2021gaitgraph,cosma2021wildgait} process sequences of skeletons as spatial-temporal graphs, using models such as ST-GCN \cite{yan2018spatial} and MS-G3D \cite{msg3d}. Despite their effectiveness, these methods have a significant drawback: they are black-boxes that provide no inherent insight into their decision process. 

Spatial-temporal graphs are inherently complex, and the feature attribution methods they employ are often opaque. For instance Teepe et al. \cite{teepe2022towards} showcased a feature activation visualization at the skeleton joint level for gait recognition, but the results are arguably as hard to interpret as the underlying model.

Furthermore, spatial-temporal graphs are typically hand-crafted \cite{caetano2019skeleton,yan2018spatial,msg3d}, likely sub-optimal, and fixed throughout the datasets and tasks. Each method has its own way of computing the spatio-temporal graph \cite{yan2018spatial,msg3d} and the approach is heuristic-based by only considering human anatomy and naive joint connections. However, the spatio-temporal dynamics are task-dependent, and densely connected graphs \cite{msg3d} suffer from problems inherent in message-passing deep graph networks, such as over-smoothing \cite{chen2020measuring}.

In this paper, we propose a method to automatically simplify the spatial-temporal graph representation for gait-based gender estimation. Our model employs two models—an upstream and a downstream model—that can modify the adjacency matrix describing each individual walking instance. This approach removes the fixed nature of the graph and allows the network to sample a different graph for each input example. Our model is trainable end-to-end by using the Straight-Through Gumbel-Softmax trick \cite{jang2017categorical}. We conduct our experiments on CASIA-B for the task of gait-based gender estimation \cite{Catruna2021FromFT}, as this is a sufficiently simple task to demonstrate the effectiveness of our approach. Importantly, we show that the resulting graphs are more interpretable and qualitatively different from fixed graphs used by ST-GCN \cite{yan2018spatial} and MS-G3D \cite{msg3d}. Our resulting graph contains only edges required for correct classification \cite{chen2018learning}. Our research contributes towards making gait analysis more robust, interpretable and adaptable to specific tasks, paving the way for more reliable gait-based biometrics.

This paper makes the following contributions:

\begin{enumerate}
    \item We propose a method to simplify the spatio-temporal graph in skeleton-based gait analysis, which provides more interpretable graphs without losing downstream performance.
    
    \item We show that our method surpasses fixed graph-based methods for gender estimation, while having a spatio-temporal graph with considerably fewer connections.
    
    \item Our method alleviates the over-smoothing problem in graph neural network optimization that arises in densely-connected input graphs / deep models. Having a simplified spatio-temporal graph reduces the embedding smoothing across nodes and aids model learning.
\end{enumerate}

\section{Related Work}

\begin{figure*}[hbt!]
    \centering
    \includegraphics[width=0.85\textwidth]{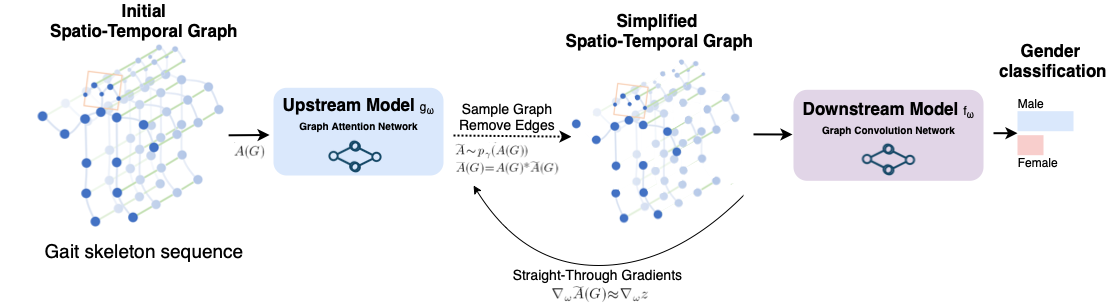}
    \caption[Caption for LOF]{High-level diagram of our proposed method. We perform gender classification of skeleton sequences by operating on a simplified spatio-temporal graph as outputted by an upstream model. The whole pipeline is trainable end-to-end by using a straight-through gradient approximation over the non-differentiable sampling operation. Skeleton sequence image is adapted from Yan et al. \cite{yan2018spatial}.
    }
    \label{fig:diagram}
\end{figure*}

\noindent \textbf{Gait Analysis.} Skeleton-based gait analysis through deep learning methods has been studied mostly in the context of gait recognition. Generally, to process skeleton sequences, some methods resort to transforming the skeleton sequences into an image and processing it with convolutional networks used in image classification. For instance, Caetano et al. \cite{caetano2019skeleton} proposed using a depth-first tree traversal order of the skeleton structure to preserve local spatial information for action recognition.  However, to better handle spatio-temporal relationships between joints, multiple methods use graph neural networks through the use of ST-GCN \cite{yan2018spatial}. Teepe et al. \cite{teepe2021gaitgraph,teepe2022towards} used an ST-GCN for processing gait sequences and obtained good results on CASIA-B \cite{CASIA:Yu}. Moreover, they showed that the model can provide insights into its decision process through activation maps. Cosma et al. \cite{cosma2021wildgait} used an ST-GCN to learn discriminative gait features in a self-supervised manner. Catruna et al. \cite{catruna2023gaitpt} used a hierarchical transformer that utilizes the anatomical structure of the human body to construct discriminative gait features. However, these methods operate on a fixed, hand-crafted spatio-temporal graph which might be sub-optimal across all tasks and datasets.

\noindent \textbf{Optimization through discrete structures.} Optimization of neural networks having discrete structures has been long utilized in the past using either Gumbel relaxations or a straight-through Gumbel-Softmax trick \cite{jang2017categorical,huijben2022review}. In terms of estimating the discrete structure of a graph, Franceschi et al. \cite{franceschi2019learning} proposed a meta-learning method to jointly learn the graph structure and the underlying parameters of the classification model. Qian and Manolache \cite{dae23qian} proposed a probabilistic data-driven graph rewiring method to estimate a per-example adjacency matrix that aids downstream classification. Significant improvements to the gumbel-softmax estimator have been made through methods such as I-MLE \cite{niepert2021implicit} and SIMPLE \cite{ahmed2023simple}. Furthermore, Chen et al. \cite{chen2018learning} have utilized discrete sampling for explainability by sampling only the most relevant features for a successful classification.

\section{Method}
\noindent \textbf{Background.} We explore spatial-temporal graph classification in the context of gait-based gender estimation \cite{Catruna2021FromFT} using sequences of skeletons extracted by a pretrained pose estimator network \cite{alphapose}. Given a dataset $\mathcal{D} = \{(\hat{y_i}, J_i)\}$ finding the best model $f_{\hat{\theta}}$ for the classification task corresponds to finding the optimal parameters $\hat{\theta}$ which minimize the cross-entropy loss $\mathcal{L}$ across the dataset: $\hat{\theta} = \argmin_\theta \mathbb{E}[\mathcal{L}(f_\theta(J), \hat{y})]$. 

A sequence of skeletons is defined by the tensor $J \in \mathbb{R}^{T \times N_j \times 2}$, in which $T$ is the number of frames and $N_j$ is the number of joints, each of which has 2 image-level coordinates ($x, y$). We use AlphaPose \cite{alphapose} to extract skeletons, which outputs $N_j = 17$ joints in COCO pose format \cite{lin2015microsoft}, with an added joint corresponding to the middle shoulders, totaling $N_j = 18$. The task of spatio-temporal classification implies the construction of a spatio-temporal graph using a heuristic that defines the relationship between joints across time $G = H(J)$, with vertices $V(G) = \{1, \dots N_j \times T\}$. A graph-based neural network \cite{gcn}, uses both the features for each joint (node) and the graph definition into its estimate: $y = f_\theta(J, G)$.

Usually \cite{cosma2021wildgait,teepe2021gaitgraph,teepe2022towards} the heuristic for constructing the spatio-temporal graph is based on the anatomical structure of the human body \cite{lin2015microsoft}, having the same joints connected across one time-step \cite{yan2018spatial} (i.e., left ankle at time step $t$ is connected with left ankle at time steps $t-1$ and $t+1$, etc.). Other methods, such as MS-G3D \cite{msg3d}, use a multi-scale approach, wiring joints across multiple node hops (including temporal connections) through the k-adjacency matrix. However, such approaches imply manual construction of the spatio-temporal graph, which is sub-optimal as it introduces inductive biases in the model. Moreover, each classification task might require different properties of the graph to handle either fine-grained motions or longer temporal interactions. Having a densely connected spatio-temporal graph (similar to \cite{msg3d}) might induce optimization issues through over-smoothing \cite{chen2020measuring}, a problem inherent in message-passing graph neural networks.

\noindent \textbf{Simplifying spatio-temporal graphs.} We propose using an additional upstream model that can adaptively rewire the spatio-temporal graph based on the training signal, similar to the approach of Qian and Manolache. \cite{dae23qian}. Following the formalization from \cite{dae23qian}, let $\mathcal{U}_n$ be the set of adjacency matrices of graphs on $n$ nodes. We define the upstream model as $g_\omega: \mathcal{U}_{N_j \times T} \times \mathbb{R}^{N_j \times T \times 2} \rightarrow \mathcal{U}_{N_j \times T}$, parameterized by $\omega$. As such, the classification output is given by  $y = f_\theta(J, g_\omega(G, J))$. The upstream model $g_\omega$ receives an initial spatio-temporal graph $G$ and the initial feature matrix $J$ and leans to output a modified graph that is best suited for the task. Concretely, $g_\omega$ outputs unnormalized edge priors $\gamma \in \mathbb{R}^{N_j \times T \times N_j \times T}$, describing the connections between each of the $N_j$ joints across all timesteps $T$. The edge priors define a probability distribution for the adjacency matrix $A(G)$: 

\begin{equation}
p_\gamma (A(G)) = \prod_{i,j}^{N_j \times T} p_{\gamma_{i,j}}(A(G)_{ij})
\end{equation}
where $p_\gamma (A(G)_{ij}  = 1) = \text{sigmoid}(\gamma_{ij})$, and $p_\gamma (A(G)_{ij}  = 0) = 1 - \text{sigmoid}(\gamma_{ij})$. In our work, we modify the initial adjacency matrix $A(G)$ by only removing its edges, without adding any additional connections, unlike Qian and Manolache \cite{dae23qian}. As such, the final adjacency matrix $\bar{A}$ is given by sampling $\tilde{A} \sim p_\gamma (A(G))$ and multiplying with the original adjacency matrix: $\bar{A}(G) = A(G) * \tilde{A}(G)$.

The sampling operation $\tilde{A} \sim p_\gamma (A(G))$ is not directly differentiable. Consequently, we employ the Straight-Through Gumbel-Softmax trick \cite{jang2017categorical} to estimate its gradients.

\begin{align}
z_{ij} &= (\gamma_{ij} - \log(-\beta\log(u_{ij} + \epsilon) + \epsilon)) / \tau \\
\tilde{A}(G)_{i,j} &= \text{round}(\text{sigmoid}(z_{ij}))
\end{align}
    
where $u_k \sim \text{Uniform}(0, 1)$, $\beta$ is a noise scaling parameter \cite{huijben2022review}, $\tau$ is the logit temperature and $\epsilon$ is a small numeric constant to avoid numeric issues. In our case, we fixed $\beta = 2$ and $\tau = 2$ for all experiments. In this context, $\nabla_\omega \tilde{A}(G) \approx \nabla_\omega z$. Figure \ref{fig:diagram} provides a high-level overview of our method. The model is trainable end-to-end by minimizing both $\theta$ and $\omega$: 

\begin{equation}
    \hat{\theta}, \hat{\omega} = \argmin_{\theta, \omega} \mathbb{E}[\mathcal{L}(f_\theta(J, g_\omega(G, J)), \hat{y})]
\end{equation}

\noindent \textbf{Model Architectures.} For the upstream model $g_\omega$, we incorporated local structure information using a Graph Attention Network \cite{velikovi2017graph,gatv2} and global information through a transformer model \cite{vaswani2017attention}. We added learned positional encodings for each timestep and added joint type embeddings to differentiate between joints. For the downstream model $f_\theta$ we use a simple graph convolutional network \cite{gcn}.

\noindent \textbf{Implementation details.} Experiments were performed on two NVIDIA RTX 2070 GPUs with 8GB of VRAM each. Each run was trained for 50 epochs with a batch size of 32, optimized using AdamW \cite{adam} optimizer. We used a learning rate of 0.001 with a cosine decay. The model number of parameters ranged from 82K for the smallest model to 2.9M for the largest. Training run duration ranged from 35 minutes for the smallest model to 2.3 hours for the largest. Since the dataset is severely imbalanced, we used balanced loss weights.

\begin{figure}
    \centering
    \includegraphics[width=\linewidth]{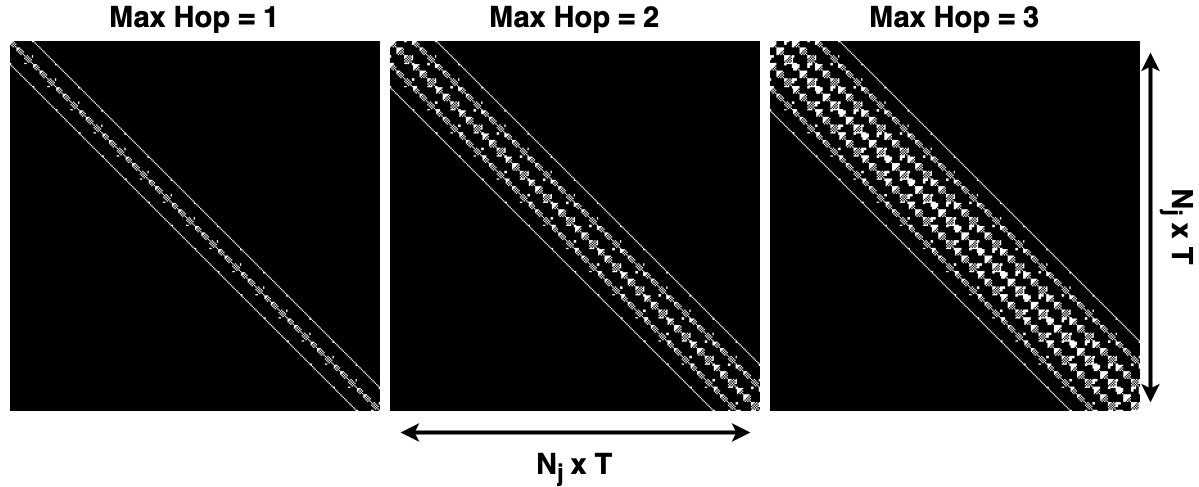}
    \caption{Adjacency matrices for the spatio-temporal graph over multiple timescales (temporal hops). The matrices are organized as blocks of $N_j \times N_j$ joints representing a timestep out of the total $T$ timesteps.}
    \label{fig:original-adj}
\end{figure}

\section{Experiments}
\begin{figure*}[hbt!]
    \centering
    \includegraphics[width=\textwidth]{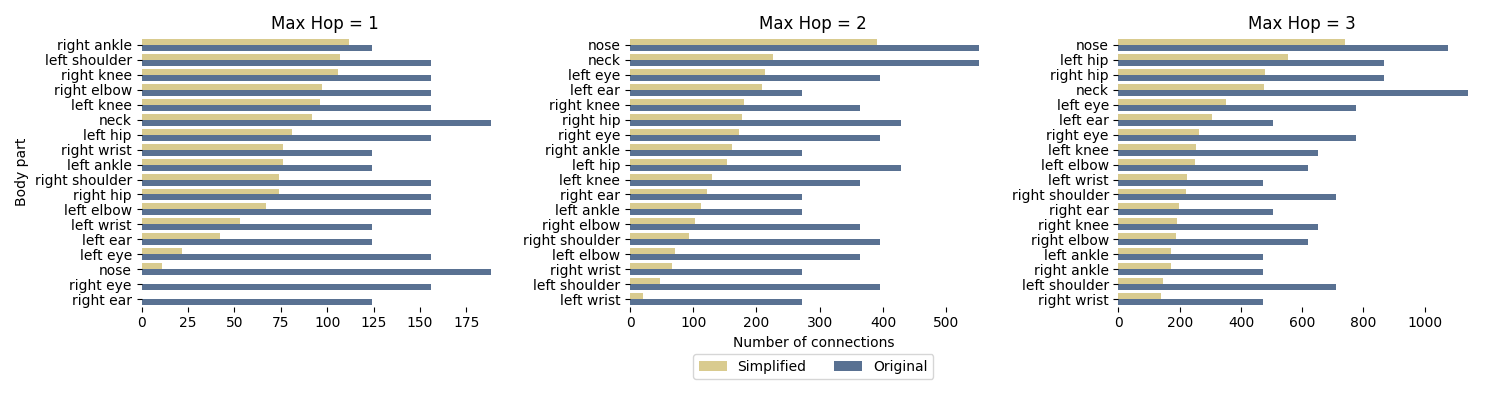}
    \caption{Number of connections per joints after graph simplification, over multiple timescales (temporal hops). With only one temporal hop, joints that are intuitively more important for gait (ankles, knees, elbows) have larger number of connections.}
    \label{fig:connections-per-node}
\end{figure*}

\begin{figure}[hbt!]
    \centering
    \includegraphics[width=1\linewidth]{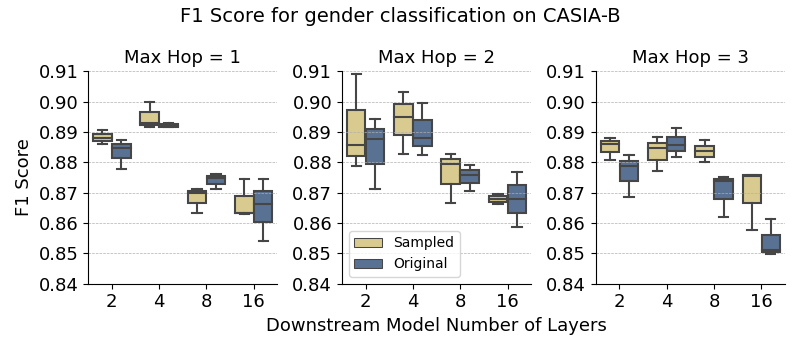}
    \caption{Results for gender classification on CASIA-B. Our method surpasses the model using a fixed initial spatio-temporal graph, even though there are fewer connections between joints.}
    \label{fig:f1-results}
\end{figure}

\begin{figure}[hbt!]
    \centering
    \includegraphics[width=1\linewidth]{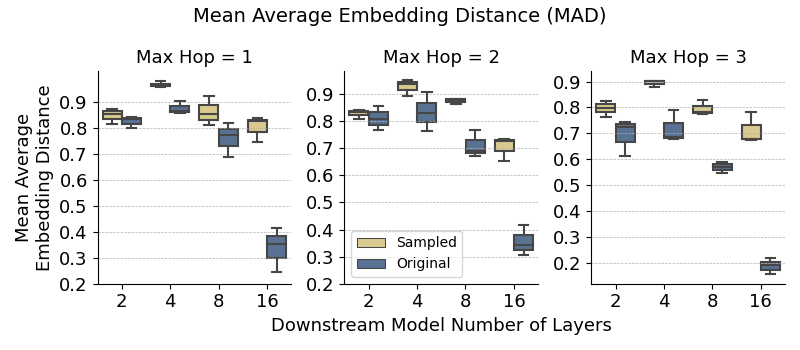}
    \caption{Results for Mean Average (Embedding) Distance for each trained model. Higher distance means less over-smoothing in the graph computation. With our method, the graph representation smoothness is greatly alleviated for deeper models / higher number of temporal hops, which aids learning.}
    \label{fig:mad-results}
\end{figure}




\noindent \textbf{Training and evaluation dataset.} In our experiments, we used the CASIA-B \cite{CASIA:Yu} gait dataset, one of the most popular datasets in the field. CASIA-B is comprised of 124 subjects walking in a straight line indoors, across 11 viewpoints. Subjects walk in three variations - normal walking (NM), clothing change (CL) and carry conditions (BG). Since CASIA-B does not contain gender annotation, we manually annotate the apparent gender of the subjects in the dataset by inspecting the raw videos. Consequently, the subjects in CASIA-B are split into 92 males and 32 females. We keep 92 individuals for training and the rest of 32 for testing. We keep the validation split balanced by keeping 16 males and 16 females.

\noindent \textbf{Experimental setup.} We study our method under multiple experimental scenarios. We are interested in quantifying downstream gender classification performance and model over-smoothing under various degrees of complexity of the downstream model. Consequently, we train downstream models having 2, 4, 8 and 16 convolutional layers. As Chen et al. \cite{chen2020measuring} pointed out, deeper models exhibit more smoothness, negatively impacting optimization. Furthermore, we study our method under three levels of temporal connectivity of the initial spatio-temporal graph. Using k-adjacency (matrix powers), similar to Liu et al. \cite{msg3d}, we compute a more dense adjacency matrix. We used $k \in \{1, 2, 3\}$, corresponding to 1, 2 and 3 maximum temporal hops, respectively. Figure \ref{fig:original-adj} showcases the adjacency matrices for the three temporal hops in our setup. Each experiment is run 3 times and we report aggregated results.

To quantify the over-smoothing problem, we use the Mean Average Distance (MAD) metric proposed by Chen et al. \cite{chen2020measuring}. The MAD metric is computed as the average pairwise cosine distance of the final node embeddings before the classification layer. Considering $\textbf{H}$ to be the embedding matrix corresponding to each node, MAD is defined as:

\begin{align}
    D_{ij} &= 1 - \frac{H_i \cdot H_j}{|H_i| \cdot |H_j|} \\
    MAD &= \frac{1}{N_j \times T \times N_j \times T} * \sum_{i,j = 0}^{N_j \times T \times N_j \times T} D_{ij}
\end{align}

Higher values of MAD imply that the node embeddings are different from one another, which translates to lower over-smoothing, while small MAD values imply a higher degree of over-smoothing.

\section{Results}

\noindent \textbf{Estimated graph is more interpretable.} Figure \ref{fig:connections-per-node} showcases the fraction of nodes kept per skeleton joint. Smaller temporal hops (i.e. max hops = 1) imply more sparsity in the spatio-temporal graph. Low sparsity increases the importance of intuitively important joints for walking: legs, hands, of a single chirality (left side / right side of the body). Leg movements (ankles, knees) are most important in gender classification, regardless of class. However, a higher number of temporal hops increases the relative importance of initially highly connected joints (i.e., nose, hips, etc.) and are less interpretable.

\noindent \textbf{Performance increases while having a simplified graph.} Figure \ref{fig:f1-results} showcases F1-scores for gender classification on CASIA-B. 
While having a much simpler spatio-temporal graph, our approach shows on par or superior performance to a fixed initial spatio-temporal graph, especially in deeper downstream models (8 and 16 number of layers) and higher temporal connectivity (max hops of 2 and 3). This is due to the alleviation of the over-smoothing problem in network training. 

\noindent \textbf{Over-smoothing is alleviated in deeper models.} Figure \ref{fig:mad-results} showcases MAD scores for each of our runs. The difference is highly visible in deeper downstream models (8 and 16 number of layers) even with only one temporal connection. The models operating on the graph with 2 and 3 temporal hops are highly affected by over-smoothing, but our model significantly reduces this issue.

\noindent \textbf{Learned simplification is superior to random edge elimination.} 
In Table \ref{tab:random} we compare with a naive method of graph simplification which presumes random edge elimination, without using the training signal for guidance. Downstream performance is severely affected with increasing number of eliminated edges. Our method is superior both in terms of F$_1$ score and MAD values.

\begin{table}[hbt!]
    \centering
    \caption{Performance comparison of our method compared with random edge elimination. We show results for a downstream model with 16 layers and 1 temporal hop.}
    \begin{tabular}{lrr}
        \textbf{Method} & \textbf{{F$_1$ $\uparrow$}} & \textbf{{MAD $\uparrow$}} \\
        \midrule
            Random Elimination 75.0\% & 0.792 {\tiny $\pm 0.063$}  & 0.191 {\tiny $\pm 0.154$}  \\
            Random Elimination 50.0\% & 0.811 {\tiny $\pm 0.073$}  & 0.264 {\tiny $\pm 0.158$}  \\
            Random Elimination 25.0\% & 0.845 {\tiny $\pm 0.079$}  & 0.507 {\tiny $\pm 0.111$}  \\
            \midrule
            Original Graph & 0.858 {\tiny $\pm 0.063$}  & 0.336 {\tiny $\pm 0.15$} \\
            \textbf{Learned Simplification (ours)} & \textbf{0.864 {\tiny $\pm 0.044$}}  & \textbf{0.803 {\tiny $\pm 0.087$}}  \\
        \end{tabular}
    \label{tab:random}
\end{table}

\section{Conclusions}
In this work, we proposed an end-to-end method to automatically simplify spatio-temporal graphs in gait classification problems. We used an upstream model that predicts a simplified graph that is then used by a downstream model for the final classification. We showed that our approach is more interpretable, has better results and is robust to over-smoothing compared to both an approach using a fixed spatio-temporal graph and compared to a naive graph simplification approach.

{\small
\bibliographystyle{ieee}
\bibliography{egbib}
}

\end{document}